\documentclass[letterpaper, 10 pt, conference]{ieeeconf}
\IEEEoverridecommandlockouts

\usepackage{cite}
\usepackage{amsmath,amssymb,amsfonts,amsthm}
\usepackage{graphicx}
\usepackage{textcomp}
\usepackage{xcolor}
\usepackage{lipsum}
\usepackage{comment}
\usepackage{multirow}
\usepackage{multicol}
\usepackage{siunitx}
\usepackage{algorithm}
\usepackage[noend]{algpseudocode}
\usepackage[normalem]{ulem}
\usepackage{subcaption}
\usepackage{tikz}
\usetikzlibrary{calc,arrows.meta}
\definecolor{mycopper}{rgb}{0.4314,0.2392,0.2039}

\def\BibTeX{{\rm B\kern-.05em{\sc i\kern-.025em b}\kern-.08em
    T\kern-.1667em\lower.7ex\hbox{E}\kern-.125emX}}

\theoremstyle{remark}

\begin{document}
\title{\huge Localization and Control of Magnetic Suture Needles in Cluttered Surgical Site with Blood and Tissue}
\author{
Will Pryor$^\dagger$,~
Yotam Barnoy$^\dagger$,~
Suraj Raval,~
Xiaolong Liu$^*$,~
Lamar Mair,~
Daniel Lerner,~
Onder Erin,~\\
Gregory D. Hager,~
Yancy Diaz-Mercado,~
Axel Krieger
\thanks{This paper  was  supported by the University of Maryland Medical Device Development Fund (MDDF) and the National Institute  of Biomedical Imaging and Bioengineering (NIBIB) of the National Institutes of Health under award number R01EB020610.}
\thanks{{$\dagger$} Will Pryor and Yotam Barnoy contributed equally to this paper.}
\thanks{* Corresponding author: Xiaolong Liu, xiaolong@jhu.edu}
\thanks{Will Pryor, Yotam Barnoy and Gregory D. Hager are with the Department of Computer Science, Johns Hopkins University, Baltimore, MD 21211 USA.}
\thanks{Suraj Raval, Daniel Lerner and Yancy Diaz-Mercado are with the Department of Mechanical Engineering, University of Maryland, College Park, MD 20742 USA.}
\thanks{Xiaolong Liu, Onder Erin and Axel Krieger are with the Department of Mechanical Engineering, Johns Hopkins University, Baltimore, MD 21211 USA.}
\thanks{Lamar O. Mair is with Weinberg Medical Physics, Inc., North Bethesda, MD 20852 USA.}
}

\maketitle

\begin{abstract}
Real-time visual localization of needles is necessary
for various surgical applications, including surgical automation and visual feedback.
In this study we investigate localization and autonomous robotic control of needles in the context of our magneto-suturing system.
Our system holds the potential for surgical manipulation with the benefit of minimal invasiveness and
reduced patient side effects.
However, the non-linear magnetic fields produce unintuitive forces and demand delicate position-based control that exceeds
the capabilities of direct human manipulation.
This makes automatic needle localization a necessity.
Our localization method combines neural network-based segmentation and classical techniques,
and we are able to consistently locate our needle with 0.73\,mm RMS error in clean environments and 2.72\,mm RMS error in challenging environments with blood and occlusion. The average localization RMS error is 2.16 mm for all environments we used in the experiments.
We combine this localization method with our closed-loop feedback control system to demonstrate
the further applicability of localization to autonomous control.
Our needle is able to follow a running suture path in (1) no blood, no tissue; (2) heavy blood, no tissue; (3) no blood, with tissue; and (4) heavy blood, with tissue environments. The tip position tracking error ranges from 2.6\,mm to 3.7\,mm RMS, opening the door towards autonomous suturing tasks.
\end{abstract}


\section{Introduction}

A surgical environment consists of many unstructured elements that make localization of surgical tools difficult.
In such conditions, the localization task is the surgical equivalent of ``finding a needle in a haystack'',
and accurate localization is a practical need for manual, tele-operated robotic, and autonomous surgical tasks.
This broad localization concept is useful for aiding the visualization of small surgical tools during both manual and robotic surgery,
as well as for surgical operations with wireless magnetic robot systems.

Magnetic robotic systems are driven by magnetic fields that allow for rapid force and torque transfer on rigid magnetic bodies.
These magnetic bodies effectively become wireless end effectors of the electromagnetic systems.
Since these magnetic end effectors can exert torques and forces,
their usage towards minimally invasive surgical operations has been explored with many applications,
including biopsy, targeted drug delivery, and capsule endoscopy~\cite{Sitti2015,Heunis2018, Liu2016}.
These wireless actuation techniques enable multi-degree-of-freedom motion on magnetic rigid bodies.
Additionally, magnetic manipulation removes the need for on-board power supplies, thus reducing surgical tool size.
Implementing magnetic suturing operations may reduce invasiveness,
increase patient comfort, reduce hospital stays, and shorten recovery times.

\begin{figure}
    \centering
    \includegraphics[width=0.95\linewidth]{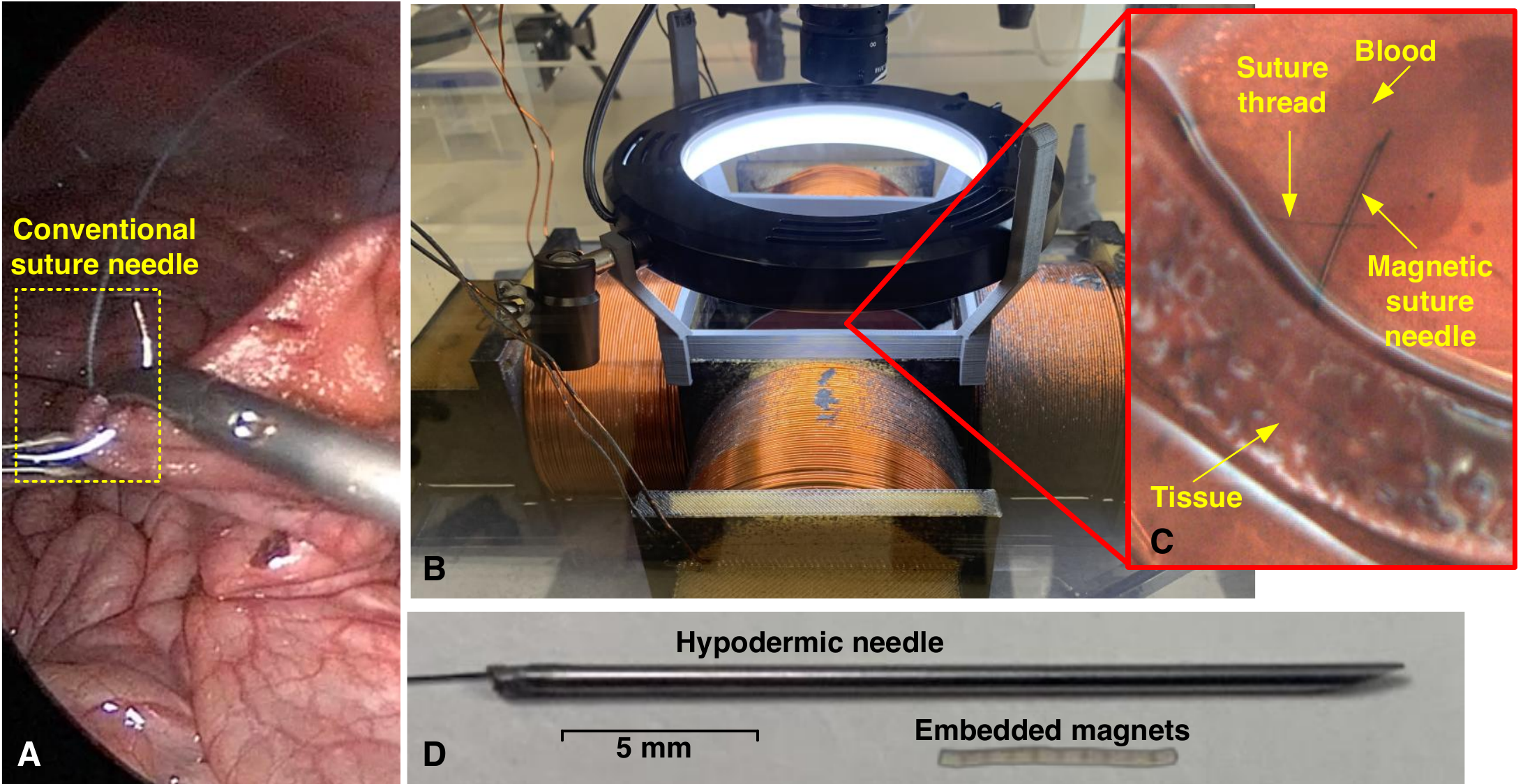}
    \caption{\small Illustration of suture tasks in a minimally invasive surgery and our proof-of-concept magnetic suture system. (A) Anastomosis of the small intestine by using a conventional suture needle. (B) The physical magnetic suture system used in this study. (C) Magnetic suture needle in a simulated cluttered surgical site with blood and tissue. (D) The magnetic suture needle used in this study.}
    \label{fig:concept}
    \vspace{-21px}
\end{figure}

The nonlinear nature of magnetic actuation systems makes direct human manipulation of magnetic robots extremely challenging.
Even worse, excessive use of magnetic force or torque could cause tissue damage or complications.
To prevent undesired outcomes and to assist surgeons with the manipulation of magnetic robots,
reliable localization algorithms combined with closed-loop feedback systems could be integrated,
moving the process closer to autonomous suturing tasks by using visual servoing via camera imaging~\cite{erin2020towards}.

For a reliable autonomous suturing task, two sub-algorithms are crucial:
(1) an accurate localization/sensing of the state of the suturing needle,
and (2) a control algorithm for calculating the desired forces and torques on the suturing needle.
Inaccurately detecting the state of the suturing needle may deceive the control algorithm and result
in undesired or unsafe forces or torques being applied.
Accurate localization and control with high closed-loop feedback rates in cluttered environments remains a challenge.
Many localization practices are limited to occlusion-free backgrounds, which oversimplify clinical scenarios~\cite{mair_magnetosuture_2020}.

Our localization goal is to track the three components of the needle's state ($x$, $y$, $\theta$) rapidly
in visually complex and cluttered environments for real-time closed-loop feedback control applications.
One known localization method makes use of feature-detection approaches to extract specific features of the needle,
such as the boundaries between different-colored sections~\cite{najafi_single-camera_2015}.
Such feature-detection is more prevalent in the broader field of surgical tool tracking~\cite{hutchison_feature_2012},
as more complex surgical tools have more easily identifiable natural features~\cite{alsheakhali_detection_2016}.
In our environment, however, the needle has few natural features.
Direct deep learning methods to predict needle position typically produce a bounding box or region containing the tool,
rather than our desired precise tool pose~\cite{nakazawa_real-time_2020, sugimori_development_2020}.
In segmentation studies,
color information is often used to separate the needle from the background~\cite{sundaresan_automated_2019, webster_image-based_2015}.
However, during surgical procedures, blood may discolor the needle,
and specular reflection from tissue may mimic metallic reflection from the needle.
In these cases, color information is insufficient to distinguish needle pixels from background pixels.
Other work uses machine learning methods,
such as random forests~\cite{chen_towards_2018} or neural networks~\cite{chang_robust_2012},
to learn segmentation models that incorporate information from neighboring pixels.
However, these works result in segmentation masks with considerable noise and
generally rely on the distinct shape of a curved needle for localization accuracy.
Numerous features in the environment incidentally resemble our straight needle and can therefore cause false detections.
The conventional design for segmentation using neural networks relies on
an encoder-decoder architecture,
where the image is first forced into an embedding space using
convolution layers, and then extracted into a segmentation mask.
This is the framework for several popular architectures such as Deeplab~\cite{chen_2017} and U-Net~\cite{ronneberger_2015}.
U-Net in particular has proven very effective at segmentation in medical imaging and has become the baseline in the field.
It applies skip connections between the encoder and decoder to
allow for different resolution levels to interact.
Due to the need for large, precise segmentation masks,
several stages of down-sampling must be applied, which tends to cause a `vanishing gradient' effect.
This is ameliorated by using more modern architectures such as
ResNets~\cite{he_deep_2015}, as in~\cite{milletari_2016}.

Applying neural networks to real-time control requires low latencies,
and thus minimal end-to-end neural network forward propagation times.
One way to speed up computation is to replace the usage of full convolutions,
which are relatively expensive to perform, in the encoder.
Dilated convolutions have been tried by~\cite{paszke_enet_2016}, though the results are lacking.
Recently, depthwise-separable convolutions with separate depth and space convolution stages
have been used by~\cite{lei_2020} and LWANet~\cite{ni_attention-guided_2020}.
These operations greatly decrease the number of required operations, driving latencies lower.
LWANet specifically utilizes MobileNetV2~\cite{sandler_mobilenet_2018},
which also makes use of depthwise-separable convolutions, as its encoder,
in addition to making use of other efficient elements such as attention fusion blocks.

In this paper, we present a novel needle localization system
feeding low-latency neural network-based segmentation into a classical localization method utilizing RANSAC~\cite{Martin1981},
thus improving on our previously presented localization~\cite{Fan2020}.
Our neural network emphasizes speed, accuracy, and a detailed output which combines
a high-resolution segmentation mask and additional information in the form of a classification output.
The proposed system maintains accurate localization in the presence of blood and tissue occlusion,
while maintaining real-time performance.
We present experiments demonstrating the improvement on prerecorded videos and in a needle control task.
The presented localization algorithm, combined with a real-time closed-loop feedback control scheme,
brings us closer to true magnetic suturing operations.

\section{Magnetic Suture System Setup and Needle Control Scheme}\label{system_scheme}

\begin{figure}[t]
    \centering
    \includegraphics[width=0.9\linewidth]{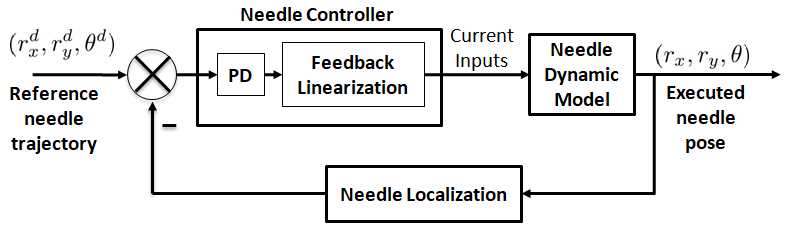}
    \caption{\small Magnetic suture needle control scheme.}
    \label{fig:closed_loop_control}
    \vspace{-15px}
\end{figure}

Our physical magnetic suture system shown in Fig.~\ref{fig:concept}B was previously presented in~\cite{mair_magnetosuture_2020,Fan2020}.
Here, we implement an updated needle design that relies on a hypodermic needle filled with NdFeB micromagnets.
Specifically, we consider the problem of controlling the motion of a 22 gauge hypodermic needle
(ID = \SI{0.413}{mm}, OD = \SI{0.7176}{mm}, length = \SI{23.5}{mm}, EXEL International 22Bx1)
with internally embedded NdFeB permanent magnets (\SI{0.3}{mm} diameter, \SI{0.5}{mm} length, 42 inserted magnets, axially magnetized).
The NdFeB magnets were sealed in place with glue.
The needle is submerged in a viscous medium with simulated blood and tissue, inside a Petri dish (diameter = \SI{85}{mm}).
Imaging was performed at 60 frames per second (FPS) using a FLIR Blackfly camera (BFS-U3-13Y3C-C) with a resolution of 1280 $\times$ 1024 pixels.
During activation, maximum currents of \SI{10}{A} were measured by current transducers (CR5411S-30 AC/DC Hall Effect Current Transducer, CR Magnetics, Inc.). All the experiments were run on a GeForce RTX 2060, with an average localization and control loop time of around 50 milliseconds. The workspace was illuminated by a ring light mounted on a custom 3D-printed adapter.
\begin{figure*}[t]
    \centering
    \includegraphics[width=0.9\linewidth]{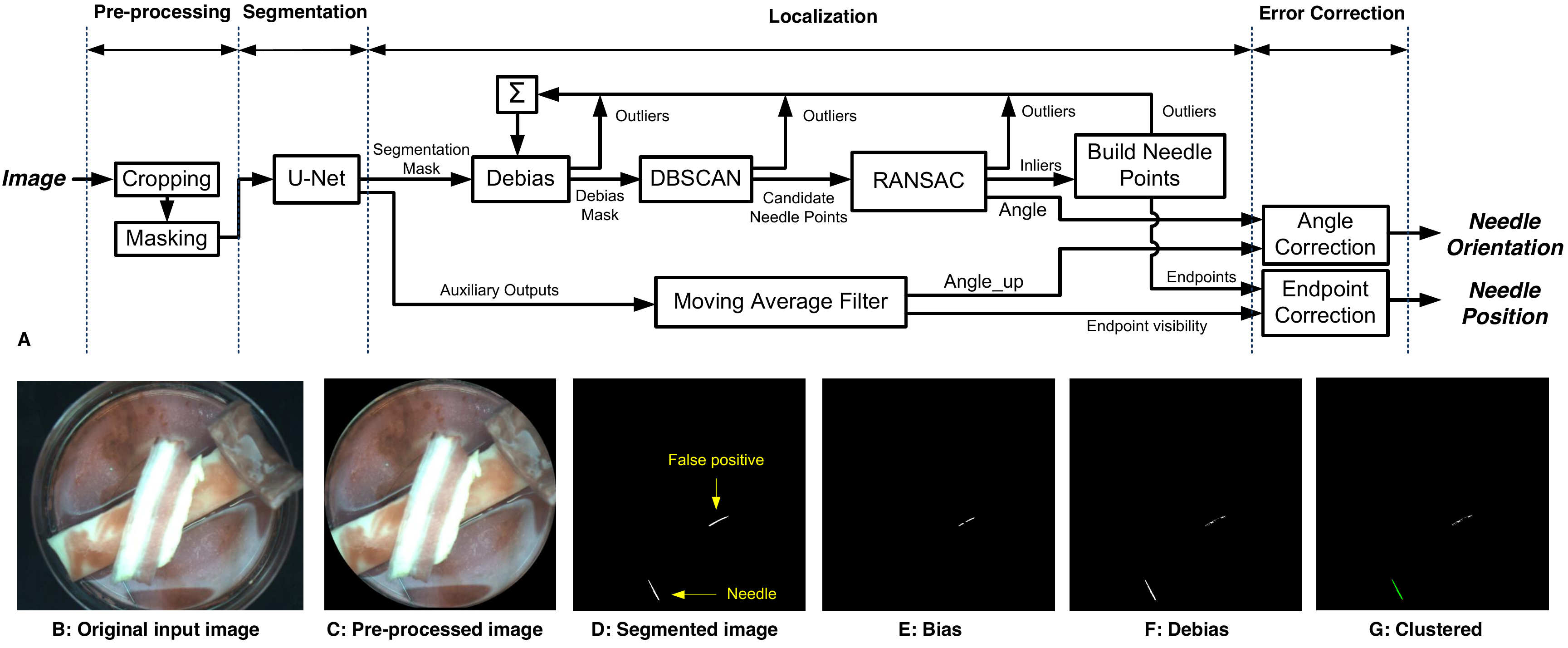}
    \caption{\small Framework of magnetic suture needle localization.}
    \label{fig:needle_loc_framework}
    \vspace{-5px}
\end{figure*}

\begin{figure*}[t]
    \centering
    \includegraphics[width=0.85\linewidth]{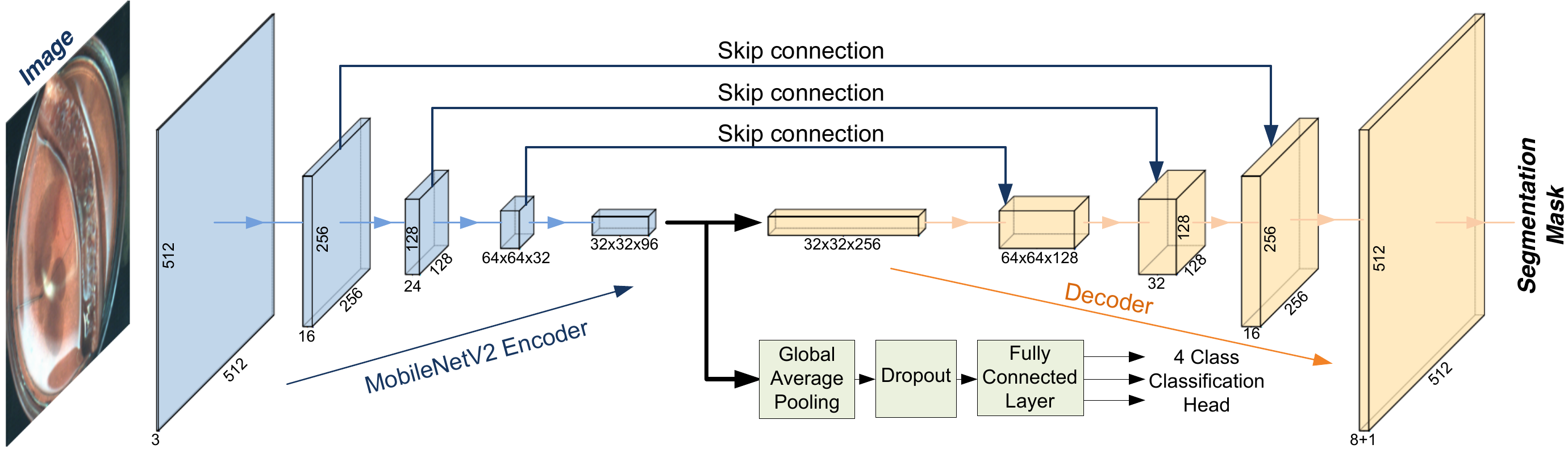}
    \caption{\small U-Net architecture for training the segmentation mask and needle tip classification.}
    \label{fig:u-net_architecture}
    \vspace{-15px}
\end{figure*}

We designed a closed-loop needle control scheme as shown in Fig.~\ref{fig:closed_loop_control},
which consists of a real-time needle localization method for dealing with cluttered surgical environments (Sec.~\ref{sec:needle_localization}),
a needle controller, and a nonholonomic model of the needle for computing the desired current inputs of the electromagnets and actuating the needle to follow a reference trajectory (Sec.~\ref{sec:needle_control}).

\section{Needle Localization}\label{sec:needle_localization}


Our needle localization aims to accurately detect the position and orientation of the 
magnetic suture needle in cluttered simulated surgical environments in real time.
Our needle localization framework (Fig.~\ref{fig:needle_loc_framework}A) consists of a pre-processing step, a neural network based segmentation step, a localization step, and an error correction step.

\subsection{Needle Segmentation}\label{subsec:needle_segmentation}
The speed and accuracy of the needle segmentation in Fig.~\ref{fig:u-net_architecture}A are equally crucial for the success of magnetic suture needle control.
Although LWANet~\cite{ni_attention-guided_2020}
targets fast test performance with its MobileNetV2\cite{sandler_mobilenet_2018} encoder,
we found that LWANet is highly sensitive to hyperparameters, slow to train,
and reduces segmentation resolution too substantially for use.
Since the MobileNetV2 architecture\cite{sandler_mobilenet_2018} by itself utilizes
depthwise-separable convolutions which require far fewer GPU operations,
we designed a U-Net~\cite{ronneberger_2015} architecture as shown in Fig.~\ref{fig:u-net_architecture} to achieve faster training speed,
better training stability, low network latency and sufficient segmentation accuracy at high resolution (512$\times$512 pixels).

Our network consists of five MobileNetV2\cite{sandler_mobilenet_2018} \textit{encoder} blocks,
each of which halves the width and height of the image,
but increases the number of features to levels prescribed by MobileNetV2 (Fig.~\ref{fig:u-net_architecture}).
Following the encoder blocks is an equal number of \textit{decoder} blocks,
each of which doubles the width and height while decreasing the number of features.
Each decoder block consists of 2 repetitions of \{convolution, batch normalization, rectified linear unit (ReLU)\}.
A skip connection connects each encoder block to the corresponding decoder block,
allowing the decoder to consider information from multiple resolution levels.
The final layer adds a convolution reducing the number of features to one.
After applying a sigmoid function, this yields our segmentation mask for either `needle' or `background'
(Fig.~\ref{fig:needle_loc_framework}D).

In order to perform suturing tasks, we need the full needle orientation, and to robustly localize the needle during partial occlusion we need an indication of whether the needle is obscured on one or both of its ends.
Therefore, we require both the segmentation mask and some additional bits of data from the neural network.
We could train a separate neural network for this classification task,
but that would entail running two neural networks on each frame,
increasing our latency.
Additionally, it's likely that information that is useful for segmentation will also have use for classification.
We therefore connect the output of the encoder to an additional classification `head'.
We use global average pooling to interpret each feature by itself,
and then add a fully connected layer with a sigmoid per output.
We train the network to produce additional bits of information about the needle:
\textit{angle\_up}, \textit{angle\_left}, \textit{tip\_visible}, and \textit{tail\_visible}. The use of these values is described in Sec.~\ref{subsec:error_correction}.

We train the segmentation head of the network using dice loss\cite{sudre_2017},
which emphasizes small classes (such as our needle) relative to the whole image,
and binary cross-entropy loss for the classification head.
The two are combined with equal weights into one loss.
We used a learning rate of 0.0001 and a Root Mean Square Propagation (RMSProp) optimizer running for 20 epochs.

\subsection{Pre-processing}\label{subsec:preprocessing}
Images are collected at 60 FPS
and for each sampled frame (Fig.~\ref{fig:needle_loc_framework}B),
the image is cropped to $1024\times1024$ pixels (Fig.~\ref{fig:needle_loc_framework}C),
which is the approximate size of the Petri dish in our sample dataset (see Sec.~\ref{subsec:dataset}).
The image is then masked by a circle with the actual radius of the dish.
Both the imaging cropping and masking are centered at the dish center, which is manually calibrated together with the radius.
Images are then downsampled to $512\times512$ pixels, and normalized per channel of RGB to a mean of ($0.485, 0.456, 0.406$) and a standard deviation of ($0.229, 0.224, 0.225$), for compatibility with pretrained weights distributed by PyTorch \cite{NEURIPS2019_9015}.
These weights are the result of training a MobileNetV2 network~\cite{sandler_mobilenet_2018} on the ImageNet dataset.

\subsection{Localization}

The needle localization method begins by running the preprocessed image through the segmentation network. We then apply a \textit{debiasing} step to correct detectable false negatives.
The segmentation's most common failure mode is to falsely identify a linear section of an obstacle boundary as a needle.
Such false positives are consistent in location throughout a single trial of the needle control and are typically shorter than a full needle.
When the needle is fully visible, the localization method successfully identifies it even when a false needle is also present because the longest needle-like object is identified.
However, when the needle is partially occluded (as shown in Fig.~\ref{fig:needle_loc_framework}B),
the false needle appears longer than the unoccluded portion of the real needle and the false needle is incorrectly identified,
as shown in Fig.~\ref{fig:needle_loc_framework}D. To address this, we learn a \textit{bias}, computed as the pixel-wise exponential moving average of the false positive rate, as illustrated in Fig.~\ref{fig:needle_loc_framework}E.
Pixels where the bias term is greater than 0.1 are subtracted from the segmentation mask.

\setlength{\textfloatsep}{5pt}
\begin{algorithm}[t]
\caption{Construction of final set of needle points}
\label{algm:1}
\begin{algorithmic}[1]
    \State Sort clusters in decreasing order by number of points
    \State $needle\_points$ := $largest\_cluster$

    \For{$cluster$ sets}
        \If{$cluster$ $\neq$ $largest\_cluster$}
            \State $d := max(||x_c - x_n||),~\forall x_c \in cluster,~\forall x_n \in needle\_points$
            \If{$d < 1.1\times needle\_length$}
                \State Append $cluster$ to $needle\_points$
            \EndIf
        \EndIf
    \EndFor
    \State \Return $needle\_points$
\end{algorithmic}
\end{algorithm}

The points in the debiased segmentation mask (Fig.~\ref{fig:needle_loc_framework}F) are clustered using the Density-Based Spatial Clustering of Applications with Noise (DBSCAN)~\cite{Martin1996} clustering algorithm.
Clusters with too few points are discarded as outliers, as are points identified as outliers by DBSCAN.
We compute the threshold $N_{min}$ for discarding a cluster as the square of the needle width.
We observe that the segmentation mask reliably produces clusters whose width is approximately equal to the needle width,
and this threshold ensures the line detection will never receive a cluster whose longer direction is along the needle width.
If no clusters remain, localization terminates and returns \textit{No Detection}.

The set of points belonging to non-discarded clusters is passed to the random sample consensus (RANSAC) method~\cite{Martin1981} for line detection.
Points identified as outliers by RANSAC are removed from their respective clusters.
If fewer than $N_{min}$ points are identified as inliers,
the localization terminates and returns \textit{No Detection}.
Otherwise the final set of needle points is constructed according to Algorithm~\ref{algm:1}.
The result is illustrated in Fig.~\ref{fig:needle_loc_framework}G.
Finally, the two extreme points in the set of needle points along the angle returned by RANSAC are taken to be the needle endpoints.

\subsection{Error Correction}
\label{subsec:error_correction}
The localization procedure identifies the needle angle and visible endpoints, but two ambiguities remain.
First, RANSAC cannot tell the difference between the needle angle $0 \leq \theta < \pi$ and the
corresponding angle $\theta^{\prime}=\theta + \pi, \pi \leq \theta^{\prime} < 2\pi$.
Accurate needle control requires knowing the correct orientation of the needle.
Second, the identified endpoints may be the true endpoints of the needle or they may be the points at which the needle disappears under an obstacle.
Proper occlusion handling requires resolving this ambiguity. We resolve both ambiguities using the outputs from the segmentation network described in~\ref{subsec:needle_segmentation}.

\textit{angle\_{up}} and \textit{angle\_{left}} are used to resolve the angle ambiguity. When the detected angle is closer to $0$ or $\pi$ than to $\pm{\pi}/{2}$, \textit{angle\_{left}} is used to resolve the ambiguity, otherwise \textit{angle\_{up}} is used. While either of these values on their own are mathematically sufficient, the network's accuracy on \textit{angle\_up} suffers when the needle is near horizontal, and likewise for \textit{angle\_left} when the needle is vertical. Choosing which to consider based on the needle angle avoids using the network output in cases where it is less accurate.

\textit{tip\_visible} and \textit{tail\_visible} are used to account for occlusion. When these values indicate both endpoints are visible, the center is computed based on the average of the endpoints. However, when only one is visible, the center is computed based on the location of the visible endpoint, the predicted angle, and the known needle dimensions. This allows the needle localization to accurately identify the needle location even with significant occlusion.
In the final case, both endpoints are occluded and we compute a best guess at the needle position based on the average of the endpoints.

\section{Needle Motion Control}\label{sec:needle_control}
In this section, we address the generation of the coil currents to allow the needle to execute a desired motion plan. In particular, we design the motion of the needle in a way that would enable the needle to perform a series of suture patters, such as ligation and running sutures. Before discussing these motion patterns, we first describe how the needle's motion is affected by the coil currents.

\subsection{Nonholonomic Model}
As described in \cite{Fan2020} in detail, and included here for completeness, in this work we assume a dipole model of interaction between the needle and the external magnetic fields generated by the coils. Under the dipole model, the magnetic fields generated by the $k$\textsuperscript{th} coil at the location of the needle's center of mass, $r=[x,y]^T$ is given by
\begin{align*}
    \mathbf{B}_k = -\frac{m_k}{\delta_k^3}(\hat{r}_k - 3\hat{d}_k\hat{d}_k^T\hat{r}_k)I_k
\end{align*}
where $m_k$ is a constant related to the magnetic properties of the coil, $r_k$ is the center location of the coil, $\hat{r}_k=r_k/\|r_k\|$ is a unit vector. The vector that points from the coil's center to the needle's center of mass is $d_k=r-r_k$, the distance between the coil and the needle is $\delta_k=d_k/\|d_k\|$, $\hat{d}_k = d_k/\delta_k$ is a unit vector, and $I_k$ is the input current at the coil.

The torque and force experienced by the needle due to the magnetic fields generated by the coils is given by the cross product between the needle's magnetic moment vector ($\mathbf{M}(\theta) = Mh$, where $M$ is a constant related to the needle's magnetic properties and $h=[\cos\theta,\sin\theta]^T$ is a unit vector pointing in the direction of the tip) and the coil's magnetic field, and the gradient of the magnetic potential, respectively,
\begin{align*}
    \tau &= \sum_{k=1}^4 \mathbf{M}(\theta)\times \mathbf{B}_k,&F &= -\sum_{k=1}^4 \nabla \left(\mathbf{M}(\theta)^T\mathbf{B}_k\right).
\end{align*}
The needle's motion is assumed to be first-order due to laminar fluid flow from low Reynolds number and negligible inertia terms \cite{salehizadeh2017two}, resulting in
\begin{align*}
    \dot{r} = \tfrac{1}{c_t}F,&& \dot{\theta} = \tfrac{1}{c_r}\tau
\end{align*}
where the coefficients $c_t$ and $c_r$ map forces and torques to translational and rotational motion (obtained empirically).

We require that the needle's translational motion to be kinematically constrained along the length of the needle, ensuring the force is applied along the tip of the needle \cite{Fan2020}. The nonholomic model becomes
\begin{align*}
    \begin{bmatrix}
        \dot{r}\\
        \dot{\theta}
    \end{bmatrix}
    =\begin{bmatrix}
        hh^T\tfrac{F}{c_t}\\
        \tfrac{\tau}{c_r}
    \end{bmatrix}
    =
    \begin{bmatrix}
        \cos\theta&0\\\sin\theta&0\\0 & 1
    \end{bmatrix}
    g(r,\theta)I
    =
    \begin{bmatrix}
        \cos\theta&0\\\sin\theta&0\\0 & 1
    \end{bmatrix}
    \begin{bmatrix}
    v\\\omega
    \end{bmatrix}
\end{align*}
where needle's linear and angular velocity are given by
\begin{align*}
    \begin{bmatrix}
    v\\\omega
    \end{bmatrix}=g(r,\theta)I = \sum_{k=1}^4 g_k(r,\theta)I_k
\end{align*}
and where if $S$ is a skew-symmetric rotation matrix,
\begin{align*}
    g_k(r,\theta) = \begin{bmatrix}
    \frac{3Mm_k}{c_t}\frac{\left(2\hat{d}_k^Thh^T\hat{r}_k + \hat{r}_k^T\hat{d}_k  - 5 \left(\hat{d}_k^Th\right)^2 \hat{d}_k^T\hat{r}_k\right)}{\delta_k^4}\\
    \frac{Mm_k}{c_r}\frac{\left(h^TS\hat{r}_k-3h^TS\hat{d}_k\hat{d}_k^T\hat{r}_k\right)}{\delta_k^3}
    \end{bmatrix}.
\end{align*}
Given a desired linear and angular velocity commands $(v,\omega)$, the input currents can be found by employing a pseudo-inverse of the motion vector field relating these, i.e.,
\begin{align*}
I = g(r,\theta)^T\left(g(r,\theta)g(r,\theta)^T\right)\begin{bmatrix}
v\\\omega
\end{bmatrix}.
\end{align*}

\subsection{Tip-Tracking Control}
In this work, we are interested in accurately controlling the tip of the needle through the dish. This will be important to ensure the tip penetrates tissue at the desired location as well as avoiding inadvertently damaging surrounding tissue. To this end, we now derive a tip-tracking controller to allow the tip of the needle to track a desired reference trajectory.

Let $r_{tip}=[x_{tip},y_{tip}]^T$ be the position of the tip relative to the center position of the needle, $r$, which is provided by the localization algorithm. As the needle is rigid, the location of the tip is related to the center via
\begin{align*}
    r_{tip} = r + \frac{\ell}{2}\begin{bmatrix}
    \cos\theta\\\sin\theta
    \end{bmatrix}
\end{align*}
where $\ell$ is the length of the needle. Differentiating these expressions, we get that
\begin{align*}
    \dot{r}_{tip} = \dot{r} + \frac{\ell}{2}\begin{bmatrix}
    -\sin\theta\\\cos\theta
    \end{bmatrix}\omega
    = \begin{bmatrix}
    \cos\theta\\\sin\theta
    \end{bmatrix}v + \frac{\ell}{2}\begin{bmatrix}
    -\sin\theta\\\cos\theta
    \end{bmatrix}\omega
\end{align*}
which suggests a unique mapping
\begin{align*}
\begin{bmatrix}
v\\\omega
\end{bmatrix} = \begin{bmatrix}
\cos\theta&\sin\theta\\-\tfrac{1}{\ell/2}\sin\theta&\tfrac{1}{\ell/2}\cos\theta
\end{bmatrix}\dot{r}_{tip}.
\end{align*}
Given a differentiable desired reference path $r_{des}(t)$, we employ a PD control strategy to enable tip tracking
\begin{align*}
    \dot{r}_{tip} = k(r_{des}(t)-r_{tip})+\dot{r}_{des}(t).
\end{align*}

\subsection{Motion Design}
The reference desired path $r_{des}(t)$ is designed to test the controller and localization algorithms at different regions within the petri-dish and under various visual conditions as described in the next section. The path is designed to achieve a back-and-forth motion of the tip, e.g., as required for a running suture. The path is parameterized in time as follows
\begin{align*}
    r_{des}(t) =
    \begin{cases}
        \frac{(t-t_{i-1})}{(t_i-t_{i-1})} (q_i-q_{i-1}) + q_{i-1}, & t\in[t_{i-1},t_i)
    \end{cases}
\end{align*}
where  $q_0,\ldots,q_m$ are a sequence of points that define the shape of the path, $t_i = t_{i-1} + \|q_i-q_{i-1}\|/v_{des}$ with $t_0=0$, and $v_{des}$ is the desired speed of the reference point through the path. The derivative of the path is given by
\begin{align*}
    \dot{r}_{des}(t) =
    \begin{cases}
        v_{des} \frac{(q_i-q_{i-1})}{\|q_i-q_{i-1}\|}, & t\in[t_{i-1},t_i)
    \end{cases}
\end{align*}
We set $v_{des} = 0.2 mm/s$. We choose this speed to take into account the effects of static and dynamic friction that the needle experiences during its motion. The points $q_0,\ldots,q_m$ are selected to mimic a running suture pattern penetrating a tissue segment in the center region of the Petri dish, making three passes through the tissue.
The tissue thickness used in the experiments is around 5 mm. The reference path and tracking performance are illustrated in Fig.~\ref{fig:controller_results}.

\section{Experimental Setup}
We created a simulated surgical site in the Petri dish using fake blood (Liquid Latex Fashions Inc.) and synthetic tissues (Abdominal Tissue Plate, SynDaver Inc.) to collect datasets for training the segmentation network and test the localization method and the  needle control performance.

\begin{figure}[t]
    \centering
    \includegraphics[width=\linewidth]{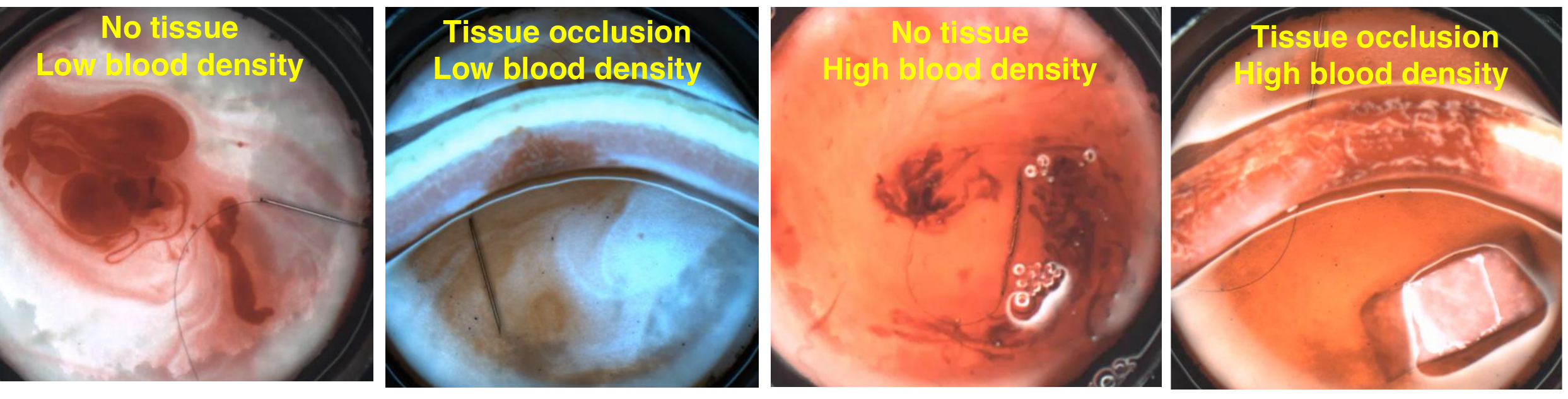}
    \caption{\small Example datasets for training segmentation network.}
    \label{fig:dataset}
\end{figure}

\subsection{Dataset Preparation for Training Segmentation Network }\label{subsec:dataset}

Various simulated surgical site conditions were set up by combining categories of different blood densities,
with or without tissue occlusion, different fluid viscosity, and different illumination conditions,
as illustrated in Fig.~\ref{fig:dataset}.
For each condition, we recorded the magnetic needle motion created using manual joystick control\cite{mair_magnetosuture_2020}.

Each training video was initially recorded at 60 FPS, and 75\% of them were then downsampled to 4 FPS due to lack of internal variation.
The needle location and orientation in each frame of the training dataset were manually annotated and converted to ground truth masks.
To maximize the variety of training data, we augmented it with random rotations and hue-saturation-luminance modifications.
In total we trained on 37,990 frames.

\subsection{Evaluation of Segmentation Network}\label{subsec:exp_setup_segmentation}
For the purpose of evaluating performance of the segmentation network and the localization method,
we classified our dataset into 4 exclusive categories:
\textbf{A}: no/low blood without tissue;
\textbf{B}: heavy blood without tissue;
\textbf{C}: no/low blood with tissue;
\textbf{D}: heavy blood with tissue.

To reliably evaluate the performance of our segmentation network, we chose to use {leave-k-out} cross-validation,
where k consists of several videos at a time.
This gives a good sense of the performance of the network in completely unseen environments. We ran our experiments with a batch size of 4 on a GeForce RTX 2080. We report average statistics across all folds of the cross-validation.


\section{Results}\label{sec:result}

\subsection{Segmentation Performance}\label{sec:segmentation_performance}

\begin{figure}[t]
    \centering
    \includegraphics[width=\linewidth,clip=true,trim=0 10 0 10]{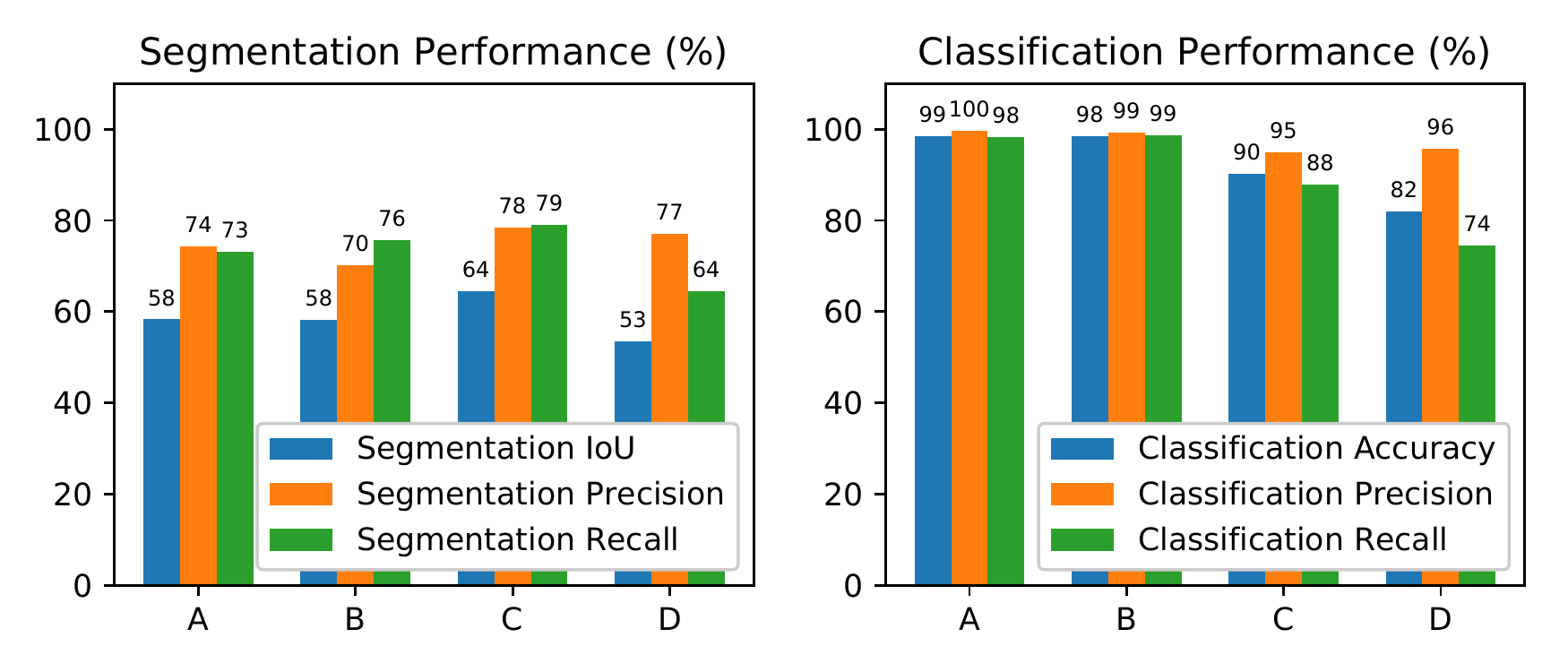}
    \caption{\small Segmentation performance and classification performance. Categories A, B, C, D are defined in Sec.~\ref{subsec:exp_setup_segmentation}. }
    \label{fig:segmentation_results}
    \vspace{-5px}
\end{figure}

Fig. \ref{fig:segmentation_results} shows the Intersection-over-Union (IoU),
precision, and recall of the segmentation component and the accuracy, precision, and recall of the classification component.
All results are averaged over leave-k-out folds,
where the network is tested on completely unseen data.

Note that while precision and recall tend to be competitive,
IoU varies between 53\% and 64\% on unseen videos, which appears low.
For example, our encoder network achieved 74.7\% IoU on the ImageNet dataset \cite{sandler_mobilenet_2018}.
However, this makes sense given the constraints of our annotation technique:
we do not have pixel-perfect mask annotations,
but rather extrapolate rectangular-needle ground-truth masks based on human location/orientation estimates and per-video tissue locations.
Additionally, our masks cannot account for shifts in tissue location over the video,
refraction through the medium changing the apparent size of the needle, heavy glare from the medium, or motion blur.
We expect some degree of inaccuracy due to the imperfections in the labels.
Furthermore, we note that our localization process is far more sensitive to recall than to precision,
since the pose detection steps effectively filter out a large degree of false positives,
but are less successful at compensating for false negatives.

\begin{figure*}[t]
    \centering
    \includegraphics[width=\linewidth,clip=true,trim=0 13 0 12]{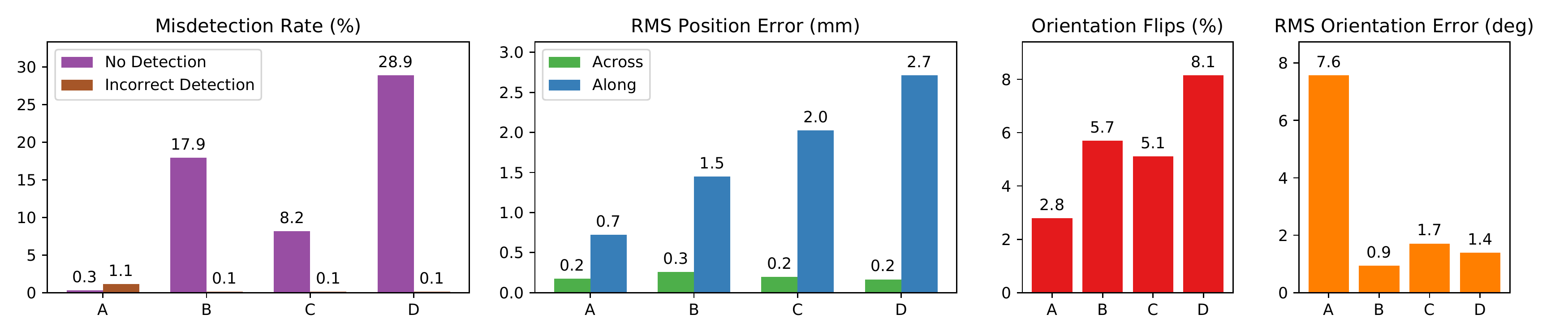}
    \caption{\small Localization performance in categories of A, B, C and D (refer to Sec.~\ref{subsec:exp_setup_segmentation}). The RMS Orientation Error represents the error after orientation flips are corrected.}
    \label{fig:localization_results}
    \vspace{-15px}
\end{figure*}

\begin{figure}[t]
    \centering
    \includegraphics[width=\linewidth]{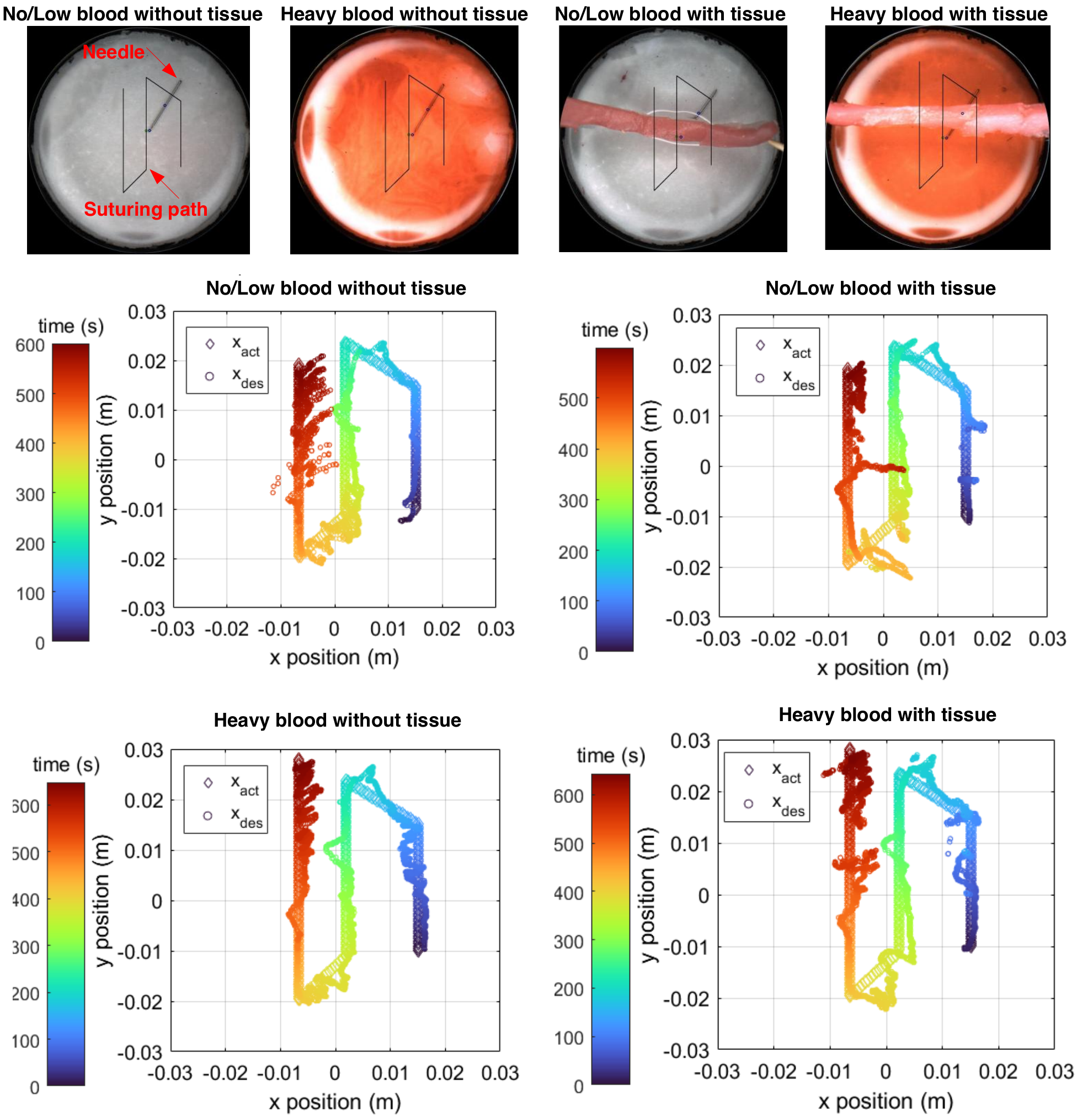}
    \caption{\small Experimental results of needle tip-tracking control performing a running suture path using the proposed localization algorithm.}
    \label{fig:controller_results}
\end{figure}
The difference between categories A and B (see Sec.~\ref{subsec:exp_setup_segmentation}) is within 5\% on all three segmentation metrics,
suggesting the network is well-equipped to deal with occlusion.
Unexpectedly, the network performs better in category C,
with occlusion but no blood, than in category A.
This may be due to imbalance in the training data,
with categories A and B each representing approximately 10\% of the training data and categories C and D each representing approximately 40\%.
However, in category D, performance is significantly lower, which is expected due to the difficulty of combining both blood and occlusion.

Classification performance on the needle angle and tip/tail visibility outputs is shown in Fig.~\ref{fig:segmentation_results}.
Categories A and B show highly similar results with near 100\% accuracy.
This suggests that the network successfully learns to identify the tip of the needle.
In contrast with segmentation, the results are lower for categories C and D.
This may indicate that the presence of blood impedes the classification's ability to identify the tip, as that information is required for all four classification outputs.

\subsection{Localization Performance}\label{sec:localization_performance}

Fig. \ref{fig:localization_results} shows the performance of the localization in the four testing categories. The misdetection rate is divided into \textit{No Detection} and \textit{Incorrect Detection}. Both errors can be corrected for rather easily using filtering over time.

The \textit{Incorrect Detection} rate is negligible in all but the first category,
where a single video in the dataset contributes 86\% of the incorrect detections.
In this video, which is not well represented in the training data, the needle moves to the very edge to the dish, causing it to be difficult to detect even for humans. If this case is excluded the incorrect detection rate is only 0.22\%, similar to the other cases.
The \textit{No Detection} rate is negligible in the first category.
In the presence of tissue (category B), the \textit{No Detection} rate increases significantly to 18\%,
and heavy blood (category C) moderately increases the rate to 7.7\%.
With heavy blood and tissue (category D) the \textit{No Detection} rate is at its highest at 29\%.
This suggests that both occlusion and heavy blood impact the detection's success rate separately,
and those errors compound when both conditions are present.

Position errors are divided into ``along needle'' and ``across needle'',
parallel and perpendicular to the needle direction, respectively.
Note that our method is significantly more accurate in the ``along needle'' direction:
the root-mean-square (RMS) error in the across direction is significantly below 1mm in all cases,
indicating a high reliability in correctly identifying a sufficient number of needle pixels to allow RANSAC to locate the needle line.
However, the RMS error along the needle has higher variance,
indicating a difficulty in determining the precise location of the needle.
As the cases increase in difficulty, the RMS error increases,
representing also an increase in the rate of classification error,
since the classification outputs are used to determine the full extent of the needle when the tip or tail is occluded.

Many orientation errors are caused by needle flips,
where the predicted orientation is off $180\deg$ from the correct orientation. This is a distinct failure mode which can be detected and mitigated by assuming the orientation has not changed more than $180\deg$ from the previous frame, and so we report the rate of orientation flips separately from the flip-corrected orientation error. Orientation flips occur in approximately 3\% of frames in the first category, 6\% with either blood or occlusion, and reach a maximum of 8\% with blood and occlusion.
As with the RMS error across the needle, this likely indicates the network's failure to identify the needle tip in more difficult cases.
The RMS orientation error, which is computed on the flip-corrected orientation, is below 2 degrees in all categories except for the first.
The high value for the first category is caused by the same video which caused errors in incorrect detection. Excluding this video, the RMS orientation error for category A is $0.81\deg$. Aside from this outlier, the network is highly successful at detecting orientation once flips are corrected. This again indicates that our method can successfully detect the needle line, and that the greatest source of error is determining the needle's location along that line.

We also compared the proposed needle localization method to an unpublished iteration of the method presented in our previous work \cite{Fan2020}. This method can only localize the suturing needle with no suture thread in the clean Petri dish with plain white background. Across 2,000 frames in those conditions the previous method achieves a RMS error of 6.18mm, compared to only 0.61mm with the proposed method. With the suture thread present, the previous method reports ``no detection'' in 68\% of frames compared to 0.64\% for the proposed method, and on the remaining frames achieves an RMSE error of 11.7mm compared to 0.86mm for the proposed method. It is clear that the previous method cannot localize the needle in any environment more complex than the clean, empty Petri dish.

\subsection{Motion Control Performance}
We experimentally validated the use of the proposed localization method and the controller under 4 types of environments: No/low blood without tissue, Heavy blood without tissue, No/low blood with tissue, and Heavy blood with tissue. The path tracking performance of the needle-tip for these conditions is provided in Fig. ~\ref{fig:controller_results}. RMS errors in position tracking are found to be 3.0, 2.6, 3.7, 2.9 mm for the aforementioned 4 conditions, respectively. One of the main reasons for such errors is the nonlinear Coulomb surface friction. Additionally, the dipole model used to compute the magnetic fields and control efforts is not accurate enough for the locations far from the center of the dish. We believe these two elements are the major contributors to the tracking errors we observe during the experiments.

\section{Conclusion}

We present a novel needle localization method,
based on a U-Net modified for speed and additional output combined with error-correcting classical methods,
and a corresponding control method which uses our localization method to navigate a needle in a lab environment.
While our control method is tailored to our specific magnetic suturing setup,
our needle localization method is applicable to any other similar contexts.

Our proposed needle localization method is able to localize the needle in over 70\% of frames even in the most challenging environment we considered, under the presence of blood and tissue. For this challenging situation, the localization RMS error is 2.7 mm. For white background with no blood and the tissue, the average RMS localization error is 0.73 mm. Overall average of RMS localization error is 2.17 mm. The needle motion control shows 2.6 to 3.7 mm RMS errors in the desired tip position over the course of a running suture path. The reliable localization and control method proposed in this study would pave the way towards autonomous wireless suturing tasks.


\section*{Acknowledgments}
We thank Margaret Pryor, who assisted with data labeling.

\bibliographystyle{IEEEtran}
\bibliography{main.bib}

\end{document}